\documentclass[journal]{IEEEtran}

\ifCLASSINFOpdf
  \usepackage[pdftex]{graphicx}
\else
  \usepackage[dvips]{graphicx}
\fi

\ifCLASSOPTIONcompsoc
  \usepackage[caption=false,font=normalsize,labelfont=sf,textfont=sf]{subfig}
\else
  \usepackage[caption=false,font=footnotesize]{subfig}
\fi

\usepackage{amsmath,amsfonts,amsthm,amssymb}

\usepackage{url}
\usepackage{multirow}

\begin{document}

\title{Multi-platform Version of StarCraft: Brood War in a Docker Container: Technical Report}

\author{Michal~\v{S}ustr, Jan~Mal\'{y} and~Michal~\v{C}ertick\'{y}
\thanks{Michal \v{C}ertick\'{y}, Jan~Mal\'{y} and Michal~\v{S}ustr are with the Games \& Simulations Research Group (http://gas.fel.cvut.cz/), Artificial Intelligence Center at Faculty of Electrical Engineering, Czech Technical University in Prague. Emails: {\scriptsize \tt{certicky@agents.fel.cvut.cz}, \tt{jan.maly@aic.fel.cvut.cz}, \tt{michal.sustr@aic.fel.cvut.cz}}}%
}

%

\maketitle
\begin{abstract}
	We present a dockerized version of a real-time strategy game StarCraft: Brood War, commonly used as a domain for AI research, with a pre-installed collection of AI developement tools supporting all the major types of StarCraft bots. This provides a convenient way to deploy StarCraft AIs on numerous hosts at once and across multiple platforms despite limited OS support of StarCraft. In this technical report, we describe the design of our Docker images and present a few use cases. 
\end{abstract}

%
\IEEEpeerreviewmaketitle

\section{Introduction}\label{secIntro}

Games have traditionally been used as domains for Artificial Intelligence (AI) research, since they represent a well-defined challenges with various degrees of complexity. They are also easy to understand and provide a way to compare the performance of AI algorithms and that of human players. 
After a recent success with popular games like Go \cite{silver2016mastering} and Poker \cite{moravvcik2017deepstack}, both of which can now be played by AIs at super-human skill level, the attention of researchers has turned to a more complex challenge represented by Real-Time Strategy (RTS) games. In addition to numerous academic researchers, commercial companies like Facebook and DeepMind have recently expressed an interest in using the most popular RTS game of all time, StarCraft\footnote{StarCraft and StarCraft: Brood War are trademarks of Blizzard Entertainment, Inc. in the U.S. and/or other countries.}, as a test environment for their AI research \cite{gibney2016google}.

Given the recent interest in StarCraft as a testbed for modern AI techniques, the research community often needs to be able to deploy multiple instances of the game at once and combine it with various other tools across different platforms. Currently, the research applicatons of StarCraft: Brood War are limited by the fact that the game is only supported on Microsoft Windows and Mac OS. Mass deployment of multiple instances tends to run into licensing issues given the proprietary nature of supported operating systems. There are a few attempts to address some of these problems, such as the TorchCraft ML framework \cite{synnaeve2016torchcraft}, but we believe that simple self-contained Docker containers\footnote{https://www.docker.com} with no accompanying requirements have a potential to provide a more versatile and easy-to-use soltion. 
Our implementation is publicly available at: https://github.com/Games-and-Simulations/sc-docker

\section{StarCraft: Brood War as a testbed for AI research}\label{secStarcraft}

In a typical RTS, a subgenre of strategy computer games, players need to develop an economy by gathering resources, building structures and researching technologies, train a military force and try to defeat their opponents with it. StarCraft games, placed within a well-knon science-fiction universe, represent a typical example of the RTS genre. Most successful entry in the StarCraft series is StarCraft: Brood War, released in 1998 by Blizzard Entertainment. Despite its considerable age, the game is still immensely popular thanks to its mechanics, depth and exceptionally balanced set of rules.

RTS games like StarCraft put additional demands on modern AI techniques, compared to board game such as Go and Chess. According to \cite{Survey2013} most important differences are:

\begin{itemize}
\item In RTS, each player has a small amount of time to decide the next action, before the game state changes.
\item RTS games are simultaneous as players can issue actions at the same time. On top of that, some actions are durative and may not be instantaneous.
\item RTS games are typically nondeterministic and partially observable. Sometimes, there is a chance that an action will fail or has unpredictable consequences. In most cases, each player can only see a part of the game world.
\item The complexity of RTS  is estimated to be many orders
of magnitude larger than any of the board game such as Go or Chess, both in terms of state-space size and in a number of actions available at each decision cycle.
\end{itemize}

Additional reason for the popularity of StarCraft: Brood War in the AI research community is the fact that the performance of AI algorithms can be directly compared in three StarCraft AI tournaments: SSCAIT, CIG and AIIDE. More information about the competitions and their contestants can be found in \cite{certicky2017current} and \cite{churchill2016starcraft}.
\section{Use cases}\label{secUseCases}

In this section, we provide a few use cases for the dockerized version of StarCraft: Brood War. These should be taken as a mere inspiration. The project has many different applications in situations, where a user needs to programatically run many StarCraft AI games, or when StarCraft needs to run on unsupported operating systems.


\begin{figure}[h]
	\centering
	\includegraphics[width=1\columnwidth]{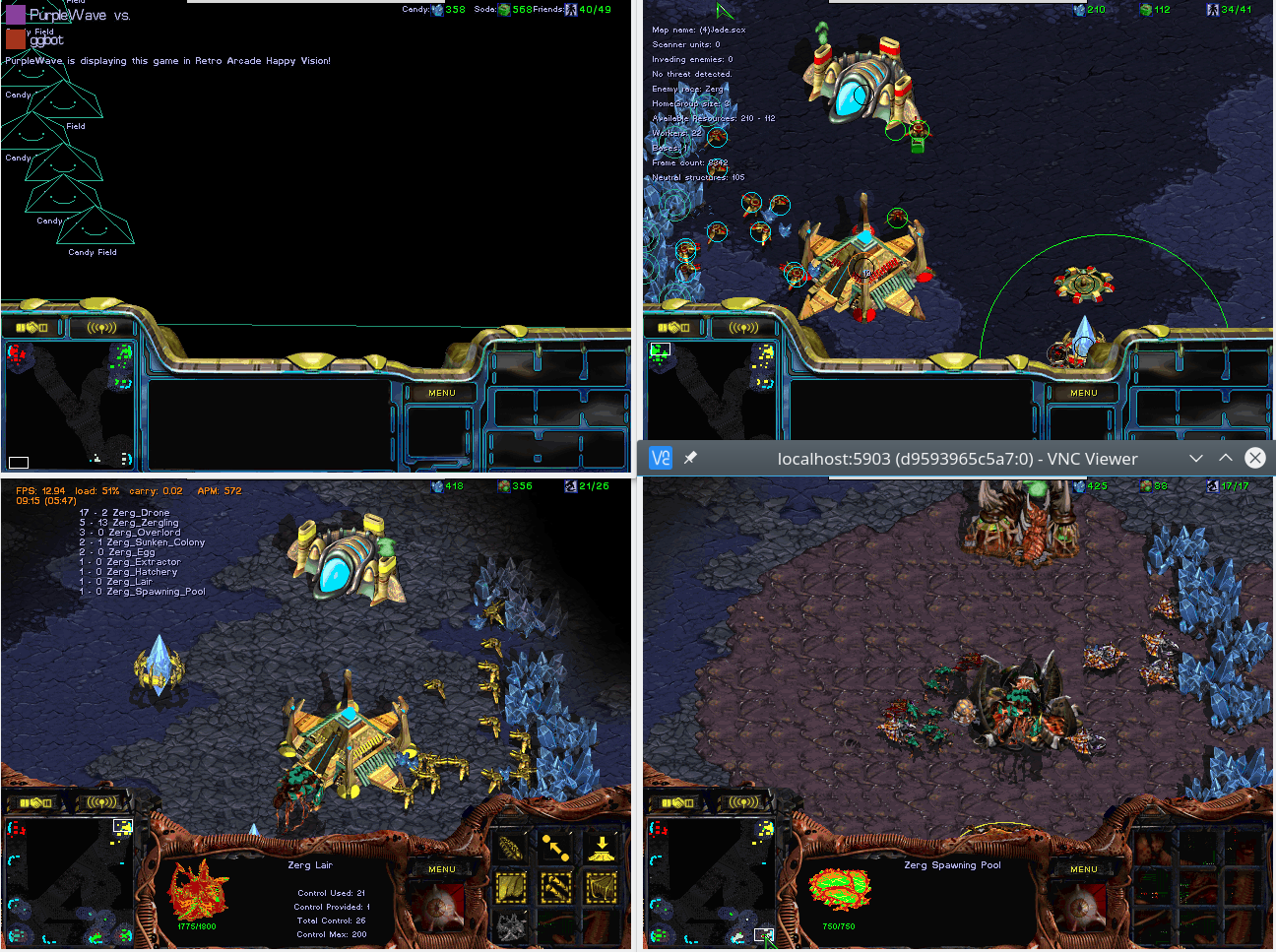}
	\caption{A LAN game of four StarCraft AI bots running on one Linux OS machine. Each bot runs in one Docker container.}
	\label{demo}
\end{figure}

\begin{itemize}
  \item \emph{Testing against different bots locally:} A bot developer can easily test his own bot against many different bots on a single machine with no need to solve their diverse requirements. Dockerized version of StarCraft contains pre-installed prerequisites of all the typical kinds of bots. User can even use provided scripts that automatically download and start the newest version of any bot from SSCAIT tournament using the tournament's API.\footnote{https://sscait.docs.apiary.io/} 
  \item \emph{Testing with varying system resources:} System resources available to each Docker container are easily configurable and can be distributed among the bots fairly. Test results should therefore be reliable and replicable. Reliability and replicability of bot's results are especially important in research. It can also help competition contestants, who can test how their bots will perform on the hardware of specific tournament. 
  \item \emph{Mass deployment for machine learning:} Running thousands of games in parallel and as quickly as possible is  vital for training ML-based bots. For example, training a bot using self-play helped master the game of Go \cite{silver2016mastering}.
     
\end{itemize}

\section{Architecture}\label{secUseCases}

The ability to extend other docker images is what makes Docker powerful. We used this feature for our containers as well. First, we created Docker image with Wine\footnote{https://www.winehq.org} (a compatibility layer allowing us to run MS Windows programs under Unix-like OS). We extended this image to add StarCraft: Brood War and a bot-coding API called BWAPI~\cite{heinermann2013bwapi}. To support Java bots, we built another container with a pre-installed Java environment. Finally, we made the last image with a collection of useful scripts allowing users to run the games easily.

Each docker container is connected to its host. One can think of this setup in terms of master/slave communication model, where docker instances are slaves and host is the master. 

\begin{figure}[h]
	\centering
	\includegraphics[width=1\columnwidth,height=10cm,keepaspectratio]{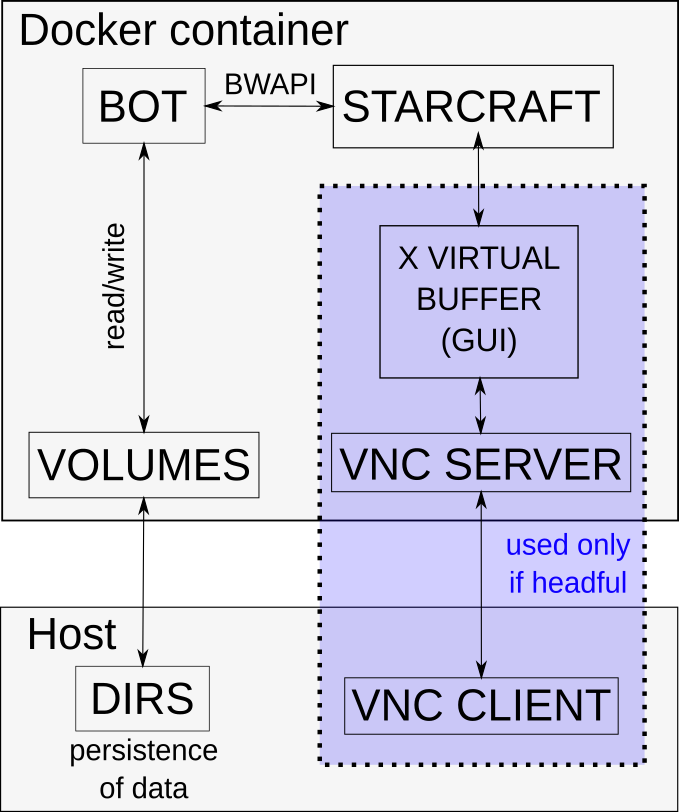}
	\caption{Communication between the Docker container and its host.}
	\label{comunication}
\end{figure}

This setup is depicted in Figure \ref{comunication}. The master contains directories (entitled DIRS in the figure) with the data read by the slaves. Directories are mapped through VOLUMES in slaves. For example, each slave can read StarCraft maps and BWAPI related data. Slaves can also write to shared directories (for logging or machine learning). Shared space is also used as cache. For example, bots share map analysis cache files created by BWTA2 library\footnote{https://bitbucket.org/auriarte/bwta2}.

Our docker container supports two modes of running StarCraft: Brood War. It is possible to run the game with a bot in so-called headless mode, which can save some computational resources by disabling GUI. This mode is useful for example for self-play. The headful mode, with the GUI enabled, is useful for debugging. In this case, the host can connect with its VNC client to VNC server running in each container as is shown in Figure \ref{comunication}.

\section{Conclusion}\label{secConclusion}


We introduced a self-contained dockerized AI research environment consisting of StarCraft: Brood War game and a collection of related AI-development tools. This provides a convenient way of massive deployment of StarCraft AIs across multiple platforms. Thanks to our scripts, it is much easier to test, train and use custom StarCraft AIs. 
Researchers and AI developers can benefit from the possibility to specify resources for container making results more reliable and replicable. We plan to add new features to our containers as well as support for OpenBW\footnote{http://www.openbw.com} and StarCraft: Remastered. We also consider extending one of AI benchmark collections represented by OpenAI Gym~\cite{1606.01540} with the StarCraft environment. 


\bibliographystyle{IEEEtran}
\bibliography{references}{} 

\end{document}